\documentclass[runningheads]{llncs}
\usepackage{graphicx}
\usepackage{amsmath}
\usepackage{amsfonts}
\usepackage{multirow}
\usepackage{float}
\usepackage{multirow}
\usepackage{color}
\usepackage[colorlinks]{hyperref}
\newcommand{\bheading}[1]{{\noindent{\textbf{#1}}}}

\usepackage{booktabs}

\usepackage{algorithmic}

\def\eg{\emph{e.g.,} }

\begin{document}

\title{Medical Aegis: Robust adversarial protectors for medical images}

\titlerunning{Medical Aegis}

\author{Qingsong Yao\inst{1} \and
Zecheng He\inst{2} \and
S. Kevin Zhou\inst{3,1}}

\institute{\
Key Lab of Intelligent Information Processing of Chinese Academy of Sciences (CAS), Institute of Computing Technology, CAS, Beijing 100190, China\\
 \email{yaoqingsong19@mails.ucas.edu.cn} \and
Princeton University \\ \and
School of Biomedical Engineering \& Suzhou Institute for Advanced Research Center for Medical Imaging, Robotics, and Analytic Computing \& LEarning (MIRACLE) University of Science and Technology of China, Suzhou 215123, China}

\authorrunning{Anonymous}

\maketitle

\begin{abstract}

Deep neural network based medical image systems are vulnerable to adversarial examples. Many defense mechanisms have been proposed in the literature, however, the existing defenses assume a passive attacker who knows little about the defense system and does not change the attack strategy according to the defense. Recent works \cite{CW_bpda,CW_ten,tramer2020adaptive} have shown that a strong adaptive attack, where an attacker is assumed to have full knowledge about the defense system, can easily bypass the existing defenses. In this paper, we propose a novel adversarial example defense system called \underline{Medical Aegis}, which boasts a two-tier protection: The first tier of \underline{Cushion} weakens the adversarial manipulation capability of an attack by removing its high-frequency components, yet posing a minimal effect on classification performance of the original image; the second tier of \underline{Shield} is inspired by the observations in our stress tests that there exist robust trails in the shallow layers of a deep network, which the adaptive attacks can hardly destruct. Thus we 
adversarially train Shield to identify and purge the adversarial features. Experimental results show that the proposed defense accurately detects adaptive attacks, with negligible overhead for model inference. 
To the best of our knowledge, Medical Aegis is \underline{the first defense} that successfully addresses the strong adaptive adversarial example attacks to medical images. 

\end{abstract}

\section{Introduction}

Deep neural networks (DNNs) are vulnerable to adversarial attacks (or attacks in short), which maliciously manipulate the network's prediction by adding human imperceptible perturbations to input images~\cite{szegedy2013intriguing}. This defect limits the application of DNNs in security-critical clinical applications~\cite{zhou2019handbook,zhou2021review}, including organ segmentation~\cite{ozbulak2019impact,daza2021towards}, disease diagnosis~\cite{paschali2018generalizability,finlayson2018adversarial}, and landmark detection~\cite{yao2020miss}. The above threats raise an urgent need for advanced approaches to help DNNs defeat the attacks and make robust decisions. There are numerous defenses developed in the literature, which are broadly categorized into two classes. (i) \textit{proactive} defense aims to reduce the adversarial noise through the use of say image filtering~\cite{jpeg} or be immune to the adversarial noise by say adversarial training~\cite{tramer2017ensemble}.
(ii) \textit{reactive} defense aims at detecting adversarial examples and then purging them. The detection is commonly done by training a classifier such as RBF-SVM~\cite{SaftyNet} and DNN~\cite{metzen2017detecting} or an anomaly detector~\cite{kde,ma2018characterizing,MAHA,li2020robust}. 

However, neither proactive nor reactive defense alone defeats adversarial attacks, especially when dealing with \textit{adaptive attacks} when the attacker has full knowledge about
the defense mechanism. For example, Backward Pass Differentiable Approximation (BPDA) \cite{CW_bpda} is able to bypass a fleet of defensive detectors. Yao \textit{et al.}~\cite{yao2020hierarchical} achieve a similar goal by designing adaptive attacks using hierarchical feature constraint (HFC). In this work, we aim to \underline{fill in this defense gap} by proposing a defense system able to address the strong adaptive adversarial attacks to medical images.


By analyzing the evolving arm race between the attacker and detector, we observe that the vulnerability of medical features is the key to empower the attacker to bypass both diagnosis networks and adversarial detectors. However, as constrained by the perturbation intensity, there must be a limit to the manipulative capability of the attacker. To confirm this, we perform a stress test as in Section \ref{sec:method} on features from different layers to quantify this capability, which aims to distort the features by attacks. Our results support the fact that deep features are vulnerable, which explains why most existing methods utilize them for reactive detection. In addition, shallow features are also affected, albeit not as significantly as deep features. \textit{This means that there are trails remaining in shallow features
of adaptive attacks, which has been largely ignored in the defense literature.}


Our goal is to defeat various conventional and adaptive adversarial attacks. \textit{Firstly},
we train a robust reactive detector, named as \underline{Shield}, with adversarial learning based on all deep features at all layers. 
The confrontation evolves into such adversarial learning as follow. On the one hand, since adaptive attacks such as HFC~\cite{yao2020miss} that take the features at various layers into consideration, the Shield improves its robustness by utilizing both shallow and deep features, hoping to exploit the trails of adaptive attacks left in various layers. 
On the other hand, the adaptive attack tries to destruct these trails, and mimics clean features in the feature space. 
\textit{Then}, we deploy a simple-yet-effective proactive step \underline{Cushion}, which weakens the adversarial attacks to arbitrarily manipulate the medical features by removing the high-frequency components, but with a minimum impact on the original image.
\textit{Finally}, we apply the proactive Cushion to help the reactive Shield win the combat more easily, resulting in a novel two-tier defense system called \underline{Medical Aegis}. The illustration of the overall framework can be found in supplemental material. 

We systematically evaluate our proposed defense on two benchmark medical image datasets, with different deep network backbones, and against both conventional and adaptive attacks. Extensive experimental results in Section \ref{sec:exp} demonstrate that: \romannumeral1) \underline{Shield} is able to defeat most of the threatening adaptive attacks. \romannumeral2) Despite denoising approaches have been proposed~\cite{jpeg} and bypassed by BPDA~\cite{CW_bpda}, \underline{Cushion} can effectively improve the robustness of the adversarial detectors against adaptive attacks.
\romannumeral3) \underline{Medical Aegis} can robustly protect a pre-trained medical image diagnosis network \textbf{with nearly 100\% accuracy}, without scarifying the original performance on disease diagnosis. To the best of our knowledge, Medical Aegis is \underline{the first defense} that successfully addresses the strong adaptive adversarial example attacks to medical images.

\section{Related Work} \label{sec:related}

\bheading{Preliminaries.} Given a clean image $x$ with its ground truth label $y \in [1,2,\ldots,Y]$, we denote a DNN classifier $h$ with pretrained parameters $\theta$; the probability $h^c_\theta(x)$ of $x$ belonging to $c$; and the logits output $l_\theta(x)$. 

\bheading{Adversarial attacks.}
Conventional white-box attacks generate adversarial examples by minimizing\footnote{In this work, we focus on targeted attack, which manipulates the network to generate the attacker-desired prediction.} the classification error between the prediction and target class $c$, while keeping the adversarial example $x^*$ within a small $\epsilon$-ball of the $L_p$-norm~\cite{PGD} centered at the clean sample $x$, \eg $\|x^* - x\|_p \leq \epsilon$, where $\epsilon$ is perturbation budget. 

These attack methods can be divided into two categories: (\romannumeral1) gradient-based approaches, \eg fast gradient sign method (FGSM) \cite{goodfellow2014explaining}, basic iterative method (BIM)~\cite{bim}, momentum iterative method (MIM)~\cite{dong2018boosting} and projected gradient descent (PGD)~\cite{PGD}, which generates $x^*$ by minimizing the loss $J(h^c_\theta(x^*))$, with $J$ often chosen as the cross-entropy loss. A typical iteration proceeds as below:
\begin{align}
   x_{t+1}^* = \Pi_\epsilon( x_t^* - \alpha \cdot \mathrm{sign}(\nabla_x J(h^c_\theta(x_t^*)))).
\label{Eq:BIM}
\end{align}
(\romannumeral2) Optimization-based methods. DeepFool~\cite{deepfool} is an $L_2$-norm attack method that moves the attacked input sample to its closest decision boundary. Furthermore, Carlini and Wagner propose the CW attack~\cite{cwattack}, which is the state-of-the-art (SOTA) white-box attack to bypass an array of neural networks and adversarial defenses~\cite{CW_ten,CW_bpda} with small perturbations. Especially, the $L_\infty$ version of CW attack can be solved by the PGD algorithm using the objective function $J_{cw}$, $l^{y_{max} \neq c}(.)$ is the maximum logits of the remaining classes, and $\kappa$ is a parameter managing the confidence:
\begin{align}
   J_{cw} = \max(l^{c}_\theta(x_t^*) - l^{{y_{max} \neq c}}_\theta(x_t^*), -\kappa).
\label{Eq:CW}
\end{align}


\bheading{Proactive defenses.} Adversarial training~\cite{tramer2017ensemble}, as one of the most representative proactive defenses with the highest robustness~\cite{dong2019benchmarking}, scarifies the model performances~\cite{tsipras2018robustness} and consumes the training time~\cite{shafahi2019adversarial}. More disturbingly, other proactive defenses like input transformation (such as JPEG compression~\cite{jpeg}) have a limited power of removing adversarial perturbations and can be bypassed by the dangerous adaptive attacks~\cite{CW_ten,tramer2020adaptive} like Backward Pass Differentiable Approximation (BPDA)~\cite{CW_bpda}. 

\bheading{Reactive defenses.} Several emerging works shed light on the intrinsic characteristic of the deep adversarial representations. Some of them use learning-based approaches (\eg RBF-SVM~\cite{SaftyNet}, DNN~\cite{metzen2017detecting}) to develop a decision boundary between clean and adversarial examples in the feature subspace. Another line of research is anomaly-detection: Feinman et al.~\cite{kde} models the clean distribution with kernel density estimation (KD). Ma et al.~\cite{ma2018characterizing} characterizes the dimensional properties of the adversarial features by local intrinsic dimensionality (LID). Lee et al.~\cite{MAHA} measures the degree of the outlier by a Mahalanobis distance based confidence score (MAHA). Sadly, all of these reactive defenses have been bypassed by adaptive attacks~\cite{yao2020hierarchical}.

\begin{table*}[htp]
\caption{Comparison of the vulnerability of the medical features from different layers, as well as using the Cushion (we use JPEG compression~\cite{tvm}) or not. We calculate the mean values of the features from different layers of ResNet-50. The relative changes of the values are reported in parentheses, bigger change indicates greater vulnerability. If the Cushion is applied, the attacker equips BIM~\cite{bim} with BPDA~\cite{CW_bpda} accordingly. The stress test uses perturbations under the constraint $L_\infty=4/256$.}
\vspace{-0.1cm}
\centering \scriptsize
\begin{tabular}{l|rrrr|r}
\bottomrule \hline
\multirow{2}{*}{Funduscopy} & \multicolumn{4}{c|}{Number of residual block }  & \multirow{2}{*}{Penultimate} \\ \cline{2-5} 
 & 1$^{st}$       & 4$^{th}$            & 10$^{th}$      & 15$^{th}$  &    \\
\hline
Clean      & .124(100\%)    & .101(100\%)     & .070(100\%)    & .233(100\%) & .425(100\%) \\
\hline
No JPEG ($\uparrow$) & .133(+7.14\%)    & .133(+31.7\%)    & .123(+73.4\%)    & .720(+210\%) & 2.471(+481\%)  \\
JPEG-90 ($\uparrow$) & .124(+0.30\%)    & .115(+13.5\%)      & .090(+25.8\%)    & .449(+92.3\%) & 1.273(+199\%)   \\
JPEG-70 ($\uparrow$) & .124(-0.52\%)    & .106(+4.58\%)      & .075(+5.60\%)      & .302(+29.9\%) & .654(+53.7\%)   \\
\hline
No JPEG ($\downarrow$) & .111(-10.9\%)    & .091(-9.78\%)      & .051(-28.1\%)     & .112(-51.8\%) & .087(-79.4\%)   \\
JPEG-90 ($\downarrow$) & .113(-9.40\%)    & .097(-3.79\%)      & .059(-16.9\%)     & .132(-44.3\%) & .103(-75.7\%)   \\
JPEG-70 ($\downarrow$) & .115(-7.80\%)    & .100(-0.69\%)      & .065(-8.99\%)     & .165(-29.2\%) & .188(-55.7\%)    \\
\hline \toprule
\end{tabular}
\vspace{-0.2cm}
\label{Table:stress}
\end{table*}

\section{Method}
\label{sec:method}

\bheading{A stress test} is performed to evaluate the robustness of features from different layers. Specifically, we decrease ($\downarrow$) and increase ($\uparrow$) the features as much as possible by attacks. In implementation, we replace the loss function $J$ in BIM (Eq.~\ref{Eq:BIM}) by $J_\downarrow^* = \mathrm{mean}(f^l_\theta(x))$ and $J_\uparrow^* = - \mathrm{mean}(f^l_\theta(x)),$ respectively, where $f^l_\theta(x)$ is the activation value from the $l^{th}$ layer. The comparison results shown in Table~\ref{Table:stress} demonstrate the following trait: adversarial attacks can alter the medical features from the deeper layers more drastically. Such a trait has both positive and negative implications. On the negative side, the adaptive attack can utilize this great vulnerability of deep features to compromise the detectors; on the positive side, \textit{it is possible to find some robust trails of the attacks, especially in the features of shallow layers}, which the attacks can not easily destruct.

\bheading{Cushion}
is inspired from the hypothesis proposed and validated in~\cite{ilyas2019adversarial}:  adversarial attacks compromise the DNNs by attributing the \textit{non-robust} signals, which are brittle, of high-frequency, imperceptible and incomprehensible to humans. Despite that using denoising approaches~\cite{jpeg} alone can not defeat BPDA~\cite{CW_bpda}, we believe that removing some of the high-frequency and imperceptible signals can prevent the adaptive attacks from manipulating medical features. To evaluate our hypothesis, we continue to perform the stress test introduced above. Differently, we apply the Cushion operation before generating medical features.

We leverage the pre-processing step called Cushion denoted by $\mathcal{C}$, which removes high-frequency components of an input image $\hat{x} = C_\lambda(x),$
where $\lambda$ is the hyper-parameter controlling the denoising degree. In this paper, we choose JPEG compression~\cite{jpeg} as the Cushion protector $\mathcal{C}$, in which $\lambda$ is the compression quality. In the ablation study, we investigate other Cushion choices too.
As shown in Table~\ref{Table:stress}, the Cushion successfully reduces the attacker's ability to manipulate medical features. Also, stronger denoising brings more robustness of medical deep representations, which is exactly what we need.

\bheading{Shield.} \label{sec:shield}
We consider a white-box confrontation between the adversarial detectors and adaptive attacks, where both sides are fully aware of the opponent's strategy and parameters. The detector is trained to identify adversarial examples generated by adaptive attacks in feature space, while the adaptive attack tries to bypass the detector by directly reducing their prediction scores in gradient back-propagation. Naturally, we perform \textit{adversarial training in the feature space} to see which side can finally win this arm race. 

Specifically, given the input $x$, the classifier $h_\theta(x)$ predicts the class $y$ and records the features $f_\theta(x)$ \textit{from shallow to deep layers}, which presents a stark contrast with existing defense approaches that mostly exploit deep features. Then, for each class $y \in [1,\ldots,Y]$, we deploy a separate DNN-based detector $\mathcal{S}$ with parameter $\phi$ that takes the features $f_\theta(x)$ as input and predicts the logit $S_\phi[f_\theta(x)]$ and the probability $\sigma(S_\phi[f_\theta(x)])$ of clean example, where $\sigma(\cdot)$ denotes the sigmoid function.

In the training phase of the detector $\mathcal{S}$, the attacker samples $\hat{x}$ from other classes and tries to bypass $\mathcal{S}$ by increasing the logit $S_\phi[f_\theta(x)]$. In practice, we use $L_\infty$-norm PGD~\cite{PGD} to generate adversarial examples $\hat{x}_s^*$ by iteratively optimizing the loss function  $J_{shield}(x) = - S_\phi[f_\theta(x)]$. The adversarial loss $L_{s}^{adv}$ is defined:
\begin{align}
   L_{s}^{adv} = \log\{\sigma(S_\phi[f_\theta(x)])\} + \log\{1 - \sigma(S_\phi[f_\theta(\hat{x}_s^*)])\}.
\label{Eq:at_shield_defense}
\end{align}
Optimized by $L_{s}^{adv}$, the detectors distinguish clean features $f_\theta(x)$ (of the class $y$) from various adversarial features $f_\theta(\hat{x}_s^*)$. As demonstrated in Table~\ref{Table:stress}, the  adversarial attacks have limited power to manipulate the shallower features. Thus, detectors can learn a tight decision boundary around the robust clean features, which are hard to mimic. Therefore, we name the robust, adversarially-trained detector as \underline{Shield}. In the inference stage, the classifier predicts the class $y$ of the input $x$ and sends the features to the corresponding detector $\mathcal{S}$, which decides to accept according to the logit $S_\phi[f_\theta(x)]$.

\bheading{Medical Aegis.} \label{sec:aegis}
We aggregate the advantages of the proactive Cushion and reactive Shield in a two-tier  defense system called ``Medical Aegis''.
In particular, for input $x$, we first use the Cushion to smooth $x$. Then we predict the smoothed image $C_\lambda(x)$ as label $y$ by classifier $h$ with feature $f_\theta(C_\lambda(x))$. Next, we generate the logit $S_\phi[f_\theta(C_\lambda(x))]$. In the training stage, the attacker crafts adversarial examples $f_\theta(\hat{x}_m^*)$ by optimizing the following object function using PGD~\cite{PGD} equipped with BPDA~\cite{CW_bpda}:
   $J_{m.aegis}(x) = - S_\phi[f_\theta(C_\lambda(x))].$
For the detector $\mathcal{S}$, the adversarial loss $L_{s}^{adv}$ is defined accordingly:
\begin{align}
   L_{m}^{adv} & = \log\{\sigma(S_\phi[f_\theta(C_\lambda(x))])\} 
   + \log\{1 - \sigma(S_\phi[f_\theta(C_\lambda(\hat{x}_m^*))])\}.
\label{Eq:at_aegis_defense}
\end{align}
The inference stage is similar to the Shield except for using the Cushion in the first stage, and the final logit is computed as $S_\phi[f_\theta(C_\lambda(x))]$.

\section{Experiments} \label{sec:exp}

\subsection{Setup}
\label{Sec:setup}

\bheading{Datasets.} Following the literature~\cite{ma2018characterizing,yao2020hierarchical,finlayson2018adversarial}, we use two representative medical datasets on classification tasks: (\romannumeral1) Kaggle Fundoscopy dataset \cite{aptos} on the diabetic retinopathy (DR) classiﬁcation task, consisting of 3,663 high-resolution fundus images. Each image is labeled to one of the five levels from ‘No DR’ to ‘mid/moderate/severe/proliferate DR’. Following \cite{ma2020understanding,finlayson2018adversarial}, we conduct a binary classification experiment, which considers all of the DR degrees as the same class. (\romannumeral2) Kaggle Chest X-Ray \cite{CXR} dataset on the pneumonia classification task, which consists of 5,863 X-Ray images labeled with `Pneumonia' or `Normal'.

\bheading{Pretrained diagnosis network.} We choose the ResNet-50 \cite{resnet} and VGG-16 \cite{VGG} models pretrained using ImageNet as diagnosis network.

\bheading{Adversarial detectors.} In experiments, we choose KD~\cite{kde}, LID~\cite{ma2018characterizing}, MAHA~\cite{MAHA}, and DNN~\cite{metzen2017detecting} as baselines. The parameters for KD, LID and MAHA are set per original papers. We compute the average of anomaly scores of all activation layers for KD, LID and MAHA as the final score. For DNN, we train a classifier for each activation layer and embed these networks by summing up their logits.

\bheading{Adversarial attacks.} For \textbf{conventional attacks}, we select fast FGSM~\cite{goodfellow2014explaining}, BIM~\cite{bim}, MIM~\cite{dong2018boosting}, PGD~\cite{PGD}, diverse input method (DIM)~\cite{xie2019improving}, translation-invariant method (TIM)~\cite{dong2019evading} and CW~\cite{cwattack}. For black-box  adversarial attacks, we choose transfer attack (TF)~\cite{liu2016delving} using VGG-16 as substitute network. Here we choose $L_\infty = 16/256$ and $L_2 = 8/256$ as the constraints. Furthermore, three \textbf{adaptive attacks} are chosen: 1) Feature attack (Fea)~\cite{feature_iclr} generates adversarial features equal to a natural one. 2) Hierarchical feature constraint (HFC)~\cite{yao2020hierarchical} pushes adversarial features toward the normal distribution of the targeted class. 3) White-box attack (WB), we assume the attacker can access the parameters of the ``sheild'' and directly reduce $S_y[f_\theta(C_\lambda(\hat{x}_m^*)]$ by optimizing the following object function:$J_{wba} = S_\phi[f_\theta(C_\lambda(\hat{x}_s^*))] + J_{cw}$.
All of the adaptive attacks generate stronger perturbations under $L_\infty=16/256$  with 100 steps with random initialization. Moreover, we strengthen the WB attack by decaying the step size~\cite{croce2020reliable} in each iteration with factor 0.1 after the 50 steps, marked as (WB$_d$). When evaluating the Cushion to strengthen the detectors, we \textit{fairly equip adversarial attacks with BPDA}~\cite{CW_bpda}.

\begin{table*}[t]
\caption{The performance of detectors (with or without the Cushion) against adversarial attacks   strengthened by HFC~\cite{yao2020hierarchical} on ResNet-50.}
\centering \scriptsize
\vspace{-0.1cm}
\begin{tabular}{l|r|r|r|r|r|r|r|r}
\bottomrule \hline 
 \multirow{2}{*}{Fundoscopy}  & \multicolumn{2}{c|}{MIM  } & \multicolumn{2}{c|}{BIM } & \multicolumn{2}{c|}{PGD  } & \multicolumn{2}{c}{CW-$l_\infty$  } \\ \cline{2-9} 
        & \multicolumn{1}{c|}{AUROC}         &  \multicolumn{1}{c|}{TPR}           & \multicolumn{1}{c|}{AUROC}         & \multicolumn{1}{c|}{TPR}      & \multicolumn{1}{c|}{AUROC}         & \multicolumn{1}{c|}{TPR}       & \multicolumn{1}{c|}{AUROC}         & \multicolumn{1}{c}{TPR}       \\ 
\hline
KD        & 50.3   & 0.9  & 46.9  & 1.4   & 57.5  & 0.5   &  46.2  & 0.8  \\
KD w/ Cushion        & 75.8   & 9.3  & 71.1  & 6.9   & 75.9  & 10.7   & 81.1  & 15.8 \\
\hline
MAHA      & 1.2     & 0.0       & 0.7  & 0.0      & 5.2     & 0.0     & 0.7  & 0.0    \\
MAHA w/ Cushion     &  80.7    &  3.7       & 75.1    & 16.3      & 82.0    & 24.7     & 83.2  & 26.0   \\
\hline
LID       & 79.9    & 30.6   & 76.0   & 27.1   & 82.8  & 43.5  &  71.8   & 20.0  \\
LID w/ Cushion    & 87.6    & 64.4   & 89.6   & 68.1   & 88.0  & 69.3  & 84.8   & 51.6  \\
\hline
DNN       & 55.6     & 18.1    & 56.5    & 29.4 & 58.8   & 33.8   & 45.1     & 0.9   \\
DNN w/ Cushion       & 89.3     & 79.1    & 67.9     & 49.3    & 71.9   & 55.8   & 89.8   & 85.6  \\
\hline
Shield     & 99.0  & 97.7     & 99.0   & 97.6      & 99.2  & 97.8     & 98.9  & 97.2   \\
Medical Aegis      & 100   & 100     & 100   & 100      & 100  & 100     & 100  & 100   \\
\hline \toprule 
\end{tabular}
\vspace{-0.3cm}
\label{Table:cushion}
\end{table*}

\begin{table*}[t]
\caption{The performances of the Shield and Medical Aegis against \textbf{conventional attacks} on different datasets and backbones.  AUC scores (\%) are used as metrics.}
\centering \scriptsize
\vspace{-0.1cm}
\begin{tabular}{l|l|ccccccccccc|}
\bottomrule \hline 
Fundoscopy &  Defends & FGSM & BIM & PGD & MIM & DIM & TIM & CW-$l_\infty$ & PGD-$l_2$ & CW & TF   \\ 
\hline
\multirow{2}{*}{ResNet-50} & Shield & 98.8 & 99.3 & 99.2 & 99.3 & 99.3 & 98.6 & 98.8 & 98.4 & 97.7 & 98.0  \\
 & Medical Aegis & 100 & 100 & 100 & 100 & 100 & 99.9 & 99.8 & 99.8 & 92.3 & 100\\
\hline 
\multirow{2}{*}{VGG-16} & Shield & 99.8 & 99.8 & 99.7 & 99.8 & 99.8 & 98.7 & 98.5 & 97.9 & 97.0 & 99.7 \\
 & Medical Aegis & 100 & 100 & 100 & 100 & 100 & 100 & 99.9 & 99.9 & 93.6 & 100\\
\hline
Chest X-Ray &  Defends & FGSM & BIM & PGD & MIM & DIM & TIM & CW-$l_\infty$ & PGD-$l_2$ & CW & TF   \\ 
\hline
\multirow{2}{*}{ResNet-50} & Shield & 89.5 & 98.1 & 98.1 & 98.3 & 98.0 & 97.9 & 97.5 & 97.2 & 96.7 & 97.8  \\
 & Medical Aegis & 100 & 100 & 100 & 100 & 100 & 100 & 100 & 100 & 100 & 100 \\
\hline 
\multirow{2}{*}{VGG-16} & Shield & 100 & 99.9 & 99.9 & 99.9 & 99.9 & 99.6 & 99.0 & 98.8 & 98.5 & 99.9 \\
 & Medical Aegis & 100 & 100 & 100 & 100 & 100 & 100 & 100 & 99.6 & 98.6 & 100 \\
\hline \toprule 
\end{tabular}
\vspace{-0.1cm}
\label{Table:convential}
\end{table*}

\begin{table}[h]
\caption{The performances of the Shield and Medical Aegis against \textbf{adaptive attacks} on different datasets and backbones. We use AUC score (\%) as the metric. }
\centering \scriptsize
\vspace{-0.1cm}
\begin{tabular}{l|l|rrrr|rrrr}
\bottomrule \hline 
\multirow{2}{*}{Backbones} &  \multirow{2}{*}{Defends} & \multicolumn{4}{c|}{Fundoscopy} & \multicolumn{4}{c}{Chest X-Ray} \\ \cline{3-10}
 & &  Fea & HFC & WB & WB$_d$ &  Fea & HFC & WB & WB$_d$   \\ 
\hline
\multirow{2}{*}{ResNet-50} & Shield & 97.6 & 97.7 & 90.6 & 92.5 & 97.8 & 96.8 & 90.2 & 92.7\\
 & Medical Aegis & 100 & 99.9 & 99.1 & 98.7 & 100 & 100 & 100 & 100 \\
\hline
\multirow{2}{*}{VGG-16} & Shield & 97.5 & 98.5 & 92.0 & 92.5 & 98.8 & 99.0 & 91.0 & 91.8\\
 & Medical Aegis & 100 & 100 & 100 & 100 & 100 & 100 & 99.7 & 99.7 \\
\hline \toprule 
\end{tabular}
\vspace{-0.1cm}
\label{Table:adaptive}
\end{table}

\bheading{Metrics.} 1) \textit{True positive rate at 90\% true negative rate (TPR):} The detector will drop 10\% of the normal samples to reject attacks;  2) \textit{AUC score}.

\bheading{Training details.} We set JPEG compression with $\lambda=80$ quality as the Cushion. The Shield extracts features from the $[1^{st}, 4^{th}, 8^{th}, 14^{th}]$ residual block in ResNet-50 and the $[3^{rd}, 6^{th}, 9^{th}, 12^{th}]$ activation layer in VGG-16. To stably train the Shield, we increase the perturbation constraint of adversarial examples from $L_\infty = 1/256$ to $L_\infty = 16 / 256$ in each epoch, with 25 epochs in total. 

\subsection{Effectiveness of Cushion}

We reproduce the combat between the state-of-the-art (SOTA) adversarial detectors against the attack using HFC~\cite{yao2020hierarchical}. As in Table~\ref{Table:cushion}, the attacker can easily utilize HFC to mimic the clean features and herein completely bypass these detectors. However, after deploying the Cushion to weaken the attacker, the bypassed detectors can identify and reject some adversarial examples.

\vspace{-0.1cm}
\subsection{Effectiveness of Medical Aegis}
We test the use of Shield and Medical Aegis to defend various SOTA conventional attacks (in Table~\ref{Table:convential}) and strong adaptive attacks (in Table~\ref{Table:adaptive}) on different backbones and datasets. 
The results show that Shield can protect the diagnosis network against conventional attacks with AUC scores larger than 95\%. Even the attackers (WB \& WB$_d$) access all of the strategies and parameters of the detectors, Shield can also distinguish most attacks with AUC scores larger than 90\%, which \textbf{greatly exceeds} other detectors in Table~\ref{Table:cushion}. The fact proves that some unique characteristics of the notorious attacks have been captured by the Shield after adversarial training in the feature space. 
Surprisingly, when the Medical Aegis relies on the Cushion to weaken the attacker, \textbf{nearly all of the  adversarial attacks even with a great perturbation budget} can not defeat our defense, despite the attacker tries to bypass the Cushion with BPDA~\cite{CW_bpda} and directly reduces the logit $S_y[f_\theta(C_\lambda(\hat{x}_m^*)]$ of the Shield by optimizing $J_{m.aegis}$. 

\bheading{Limitations:} There is a performance drop for the Aegis when detecting CW attack on Fundoscopy dataset (in Table~\ref{Table:convential}). Shield, as a training-based approach, we guess, may overfit adaptive attacks, which will be improved in the future work.

\subsection{Ablation study} \label{sec:ab_study}

\bheading{Shallow vs deep features.} 
 Based on the finding confirmed in Table~\ref{Table:stress}, we hypothesize that \textit{adaptive attacks may leave more trails in shallow features than in deep one}. To explore which layer offers more help to the Shield in identifying the notarial attacks, we separately extract features from the $[3^{rd}, 6^{th}, 9^{th}, 12^{th}]$ activation layer of the pretrained VGG-16 and train a new Shield respectively. 
As shown in Table~\ref{Table:ab_layer}, the features from a shallow layer contribute most of the performances, while the detector only trained on the features from a deep layer is completely defeated by the adaptive attacks. This result exactly confirms our conjecture above. Therefore, it is encouraged to detect adaptive attacks in the shallow layers instead of deep ones. In addition, extracting features from multiple layers help the Shield achieve the best performances.

\begin{table}[t]
\centering \scriptsize
\caption{Comparison of the performances using the features from different layers of VGG-16 and different choices as the Cushions. We use AUC score (\%) as the metric.}
\vspace{-0.1cm}
\begin{tabular}{cccc|rr|rr|c|r|rrrr}
\bottomrule \hline
\multicolumn{4}{c|}{Layer Index} & \multirow{2}{*}{PGD} & \multirow{2}{*}{CW}  & \multirow{2}{*}{HFC}  & \multirow{2}{*}{WB} & \multirow{2}{*}{Cushions} & \multirow{2}{*}{Model} & \multirow{2}{*}{PGD} & \multirow{2}{*}{Fea} & \multirow{2}{*}{HFC} & \multirow{2}{*}{WB}  \\ \cline{1-4}
3 & 6 & 9 & 12 & & & & & & & & & & \\
\hline 
$\checkmark$ & $\checkmark$ & $\checkmark$ & $\checkmark$  & 100 & 93.6 & 100 & 100 & No ``Cushion'' & 99.7 & 99.7 & 97.7 & 97.7 & 92.0 \\
\hline 
$\checkmark$  & $\times$ & $\times$ & $\times$ & 100 & 87.6 & 100 & 99.9 & JPEG-90 & 99.7 & 100 & 100 & 100 & 99.6\\
$\times$ & $\checkmark$  & $\times$ & $\times$& 100 & 87.5 & 99.9 & 97.8 & JPEG-80 & 99.6 & 100 & 100 & 100 & 100\\
$\times$ & $\times$& $\checkmark$  & $\times$& 84.1 & 86.2 & 86.8 & 36.2 & JPEG-70 & 99.1 & 100 & 99.1 & 100 & 100\\
$\times$ & $\times$& $\times$& $\checkmark$  & 100 & 77.2 & 76.3 & 41.9 & Bit-3 & 99.3 & 100 & 100 & 100 & 99.3 \\
-&- &- &- &- &- &- &- & Bit-4 & 96.8 & 100 & 100 & 100 & 100 \\
-&- &- &- &- &- &- &- & TVM & 99.2 & 100  & 100 & 100 & 100\\
\hline \toprule 
\end{tabular}
\vspace{-0.3cm}
\label{Table:ab_layer}
\end{table}

\bheading{The choice of Cushion.} We choose different denoising approaches as Cushions, such as JPEG compression~\cite{jpeg} with different qualities, Bit compression (clipping the last 3 and 4 bits) and TVM~\cite{tvm}. The performances in Table~\ref{Table:ab_layer} validate that all of the denoising blocks are able to help the Shield defend the adaptive attacks with nearly 100\% accuracy. At the same time, using these Cushions affects the performance of the diagnosis network slightly.

\section{Conclusion} \label{sec:conclusion}

In this paper, we propose Medical Aegis, a robust two-tier adversarial defense system, which consists of two parts: (i) a Cushion weakens the attacker's ability to manipulate the medical features; and (ii) a Shield trained to capture the trails left by the adversarial attack in the latent feature space, especially in the shallow layers.
Abundant experiments on two representative public medical disease diagnosis datasets and different backbones validate the effectiveness of Medical Aegis to successfully defeats most of the adversarial examples with high accuracy, including the strongest adaptive attack, by aggregating the advantages of Cushion and Shield. In the future, we plan to improve the generalization of Medical Aegis on the various adversarial attacks. 
\newpage
{\small
\bibliographystyle{ieee_fullname}
\bibliography{Main}

\begin{thebibliography}{10}\itemsep=-1pt

\bibitem{CW_bpda}
Anish Athalye, Nicholas Carlini, and David Wagner.
\newblock Obfuscated gradients give a false sense of security: Circumventing
  defenses to adversarial examples.
\newblock In {\em ICLR}, 2018.

\bibitem{CW_ten}
Nicholas Carlini and David Wagner.
\newblock Adversarial examples are not easily detected: Bypassing ten detection
  methods.
\newblock In {\em Proceedings of the 10th ACM Workshop on Artificial
  Intelligence and Security}, pages 3--14, 2017.

\bibitem{cwattack}
Nicholas Carlini and David Wagner.
\newblock Towards evaluating the robustness of neural networks.
\newblock In {\em IEEE S{$\&$}P}, pages 39--57, 2017.

\bibitem{croce2020reliable}
Francesco Croce and Matthias Hein.
\newblock Reliable evaluation of adversarial robustness with an ensemble of
  diverse parameter-free attacks.
\newblock In {\em International conference on machine learning}, pages
  2206--2216. PMLR, 2020.

\bibitem{daza2021towards}
Laura Daza, Juan~C P{\'e}rez, and Pablo Arbel{\'a}ez.
\newblock Towards robust general medical image segmentation.
\newblock In {\em MICCAI}, pages 3--13. Springer, 2021.

\bibitem{dong2019benchmarking}
Yinpeng Dong, Qi-An Fu, Xiao Yang, Tianyu Pang, Hang Su, Zihao Xiao, and Jun
  Zhu.
\newblock Benchmarking adversarial robustness.
\newblock In {\em CVPR}, 2020.

\bibitem{dong2018boosting}
Yinpeng Dong, Fangzhou Liao, Tianyu Pang, Hang Su, Jun Zhu, Xiaolin Hu, and
  Jianguo Li.
\newblock Boosting adversarial attacks with momentum.
\newblock In {\em CVPR}, pages 9185--9193, 2018.

\bibitem{dong2019evading}
Yinpeng Dong, Tianyu Pang, Hang Su, and Jun Zhu.
\newblock Evading defenses to transferable adversarial examples by
  translation-invariant attacks.
\newblock In {\em CVPR}, 2019.

\bibitem{jpeg}
Gintare~Karolina Dziugaite, Zoubin Ghahramani, and Daniel~M Roy.
\newblock A study of the effect of {JPG} compression on adversarial images.
\newblock {\em arXiv:1608.00853}, 2016.

\bibitem{kde}
Reuben Feinman, Ryan~R Curtin, Saurabh Shintre, and Andrew~B Gardner.
\newblock Detecting adversarial samples from artifacts.
\newblock {\em arXiv:1703.00410}, 2017.

\bibitem{finlayson2018adversarial}
Samuel~G Finlayson, Hyung~Won Chung, Isaac~S Kohane, and Andrew~L Beam.
\newblock Adversarial attacks against medical deep learning systems.
\newblock {\em Science}, 363(6433):1287--1289, 2018.

\bibitem{goodfellow2014explaining}
Ian~J Goodfellow, Jonathon Shlens, and Christian Szegedy.
\newblock Explaining and harnessing adversarial examples.
\newblock In {\em ICLR}, 2015.

\bibitem{tvm}
Chuan Guo, Mayank Rana, Moustapha Cisse, and Laurens Van Der~Maaten.
\newblock Countering adversarial images using input transformations.
\newblock {\em arXiv:1711.00117}, 2017.

\bibitem{resnet}
Kaiming He, Xiangyu Zhang, Shaoqing Ren, and Jian Sun.
\newblock Deep residual learning for image recognition.
\newblock In {\em CVPR}, pages 770--778, 2016.

\bibitem{ilyas2019adversarial}
Andrew Ilyas, Shibani Santurkar, Dimitris Tsipras, Logan Engstrom, Brandon
  Tran, and Aleksander Madry.
\newblock Adversarial examples are not bugs, they are features.
\newblock {\em NIPS}, 2019.

\bibitem{aptos}
Kaggle.
\newblock {APTOS 2019 Blindness Detection}, 2019.
\newblock \url{https://www.kaggle.com/c/aptos2019-blindness-detection}.

\bibitem{CXR}
Kaggle.
\newblock {"Chest X-Ray Images (Pneumonia)}, 2019.
\newblock \url{https://www.kaggle.com/paultimothymooney/chest-xray-pneumonia}.

\bibitem{bim}
Alexey Kurakin, Ian Goodfellow, and Samy Bengio.
\newblock Adversarial machine learning at scale.
\newblock In {\em ICLR}, 2017.

\bibitem{MAHA}
Kimin Lee, Kibok Lee, Honglak Lee, and Jinwoo Shin.
\newblock A simple unified framework for detecting out-of-distribution samples
  and adversarial attacks.
\newblock In {\em NIPS}, pages 7167--7177, 2018.

\bibitem{li2020robust}
Xin Li and Dongxiao Zhu.
\newblock Robust detection of adversarial attacks on medical images.
\newblock In {\em ISBI}, pages 1154--1158. IEEE, 2020.

\bibitem{liu2016delving}
Yanpei Liu, Xinyun Chen, Chang Liu, and Dawn Song.
\newblock Delving into transferable adversarial examples and black-box attacks.
\newblock In {\em ICLR}, 2017.

\bibitem{SaftyNet}
Jiajun Lu, Theerasit Issaranon, and David Forsyth.
\newblock {SafetyNet}: Detecting and rejecting adversarial examples robustly.
\newblock In {\em ICCV}, Oct 2017.

\bibitem{ma2018characterizing}
Xingjun Ma, Bo Li, Yisen Wang, Sarah~M Erfani, Sudanthi Wijewickrema, Grant
  Schoenebeck, Dawn Song, Michael~E Houle, and James Bailey.
\newblock Characterizing adversarial subspaces using local intrinsic
  dimensionality.
\newblock In {\em ICLR}, 2018.

\bibitem{ma2020understanding}
Xingjun Ma, Yuhao Niu, Lin Gu, Yisen Wang, Yitian Zhao, James Bailey, and Feng
  Lu.
\newblock Understanding adversarial attacks on deep learning based medical
  image analysis systems.
\newblock {\em Pattern Recognition}, 2020.

\bibitem{PGD}
Aleksander Madry, Aleksandar Makelov, Ludwig Schmidt, Dimitris Tsipras, and
  Adrian Vladu.
\newblock Towards deep learning models resistant to adversarial attacks.
\newblock In {\em ICLR}, 2018.

\bibitem{metzen2017detecting}
Jan~Hendrik Metzen, Tim Genewein, Volker Fischer, and Bastian Bischoff.
\newblock On detecting adversarial perturbations.
\newblock In {\em ICLR}, 2017.

\bibitem{deepfool}
Seyed-Mohsen Moosavi-Dezfooli, Alhussein Fawzi, and Pascal Frossard.
\newblock {DeepFool}: a simple and accurate method to fool deep neural
  networks.
\newblock In {\em CVPR}, pages 2574--2582, 2016.

\bibitem{ozbulak2019impact}
Utku Ozbulak, Arnout Van~Messem, and Wesley De~Neve.
\newblock Impact of adversarial examples on deep learning models for biomedical
  image segmentation.
\newblock In {\em MICCAI}, pages 300--308. Springer, 2019.

\bibitem{paschali2018generalizability}
Magdalini Paschali, Sailesh Conjeti, Fernando Navarro, and Nassir Navab.
\newblock Generalizability vs. robustness: investigating medical imaging
  networks using adversarial examples.
\newblock In {\em MICCAI}, pages 493--501. Springer, 2018.

\bibitem{feature_iclr}
Sara Sabour, Yanshuai Cao, Fartash Faghri, and David~J Fleet.
\newblock Adversarial manipulation of deep representations.
\newblock In {\em IEEE S$\&$P}, 2016.

\bibitem{shafahi2019adversarial}
Ali Shafahi, Mahyar Najibi, John Dickerson, Christoph Studer, Larry~S Davis,
  Gavin Taylor, and Tom Goldstein.
\newblock Adversarial training for free!
\newblock {\em NIPS}, 2019.

\bibitem{VGG}
Karen Simonyan and Andrew Zisserman.
\newblock Very deep convolutional networks for large-scale image recognition.
\newblock In {\em ICLR}, 2015.

\bibitem{szegedy2013intriguing}
Christian Szegedy, Wojciech Zaremba, Ilya Sutskever, Joan Bruna, Dumitru Erhan,
  Ian Goodfellow, and Rob Fergus.
\newblock Intriguing properties of neural networks.
\newblock In {\em ICLR}, 2014.

\bibitem{tramer2020adaptive}
Florian Tramer, Nicholas Carlini, Wieland Brendel, and Aleksander Madry.
\newblock On adaptive attacks to adversarial example defenses.
\newblock In {\em NIPS}, 2020.

\bibitem{tramer2017ensemble}
Florian Tram{\`e}r, Alexey Kurakin, Nicolas Papernot, Ian Goodfellow, Dan
  Boneh, and Patrick McDaniel.
\newblock Ensemble adversarial training: Attacks and defenses.
\newblock In {\em ICLR}, 2018.

\bibitem{tsipras2018robustness}
Dimitris Tsipras, Shibani Santurkar, Logan Engstrom, Alexander Turner, and
  Aleksander Madry.
\newblock Robustness may be at odds with accuracy.
\newblock {\em ICLR}, 2019.

\bibitem{xie2019improving}
Cihang Xie, Zhishuai Zhang, Yuyin Zhou, Song Bai, Jianyu Wang, Zhou Ren, and
  Alan~L Yuille.
\newblock Improving transferability of adversarial examples with input
  diversity.
\newblock In {\em CVPR}, pages 2730--2739, 2019.

\bibitem{yao2020miss}
Qingsong Yao, Zecheng He, Hu Han, and S~Kevin Zhou.
\newblock Miss the point: Targeted adversarial attack on multiple landmark
  detection.
\newblock In {\em MICCAI}, 2020.

\bibitem{yao2020hierarchical}
Qingsong Yao, Zecheng He, Yefeng Zheng, and S~Kevin Zhou.
\newblock A hierarchical feature constraint to camouflage medical adversarial
  attacks.
\newblock {\em MICCAI}, 2021.

\bibitem{zhou2021review}
S~Kevin Zhou, Hayit Greenspan, Christos Davatzikos, James~S Duncan, Bram
  Van~Ginneken, Anant Madabhushi, Jerry~L Prince, Daniel Rueckert, and Ronald~M
  Summers.
\newblock A review of deep learning in medical imaging: Imaging traits,
  technology trends, case studies with progress highlights, and future
  promises.
\newblock {\em Proceedings of the IEEE}, 2021.

\bibitem{zhou2019handbook}
S~Kevin Zhou, Daniel Rueckert, and Gabor Fichtinger.
\newblock {\em Handbook of medical image computing and computer assisted
  intervention}.
\newblock Academic Press, 2019.

\end{thebibliography}
}

\end{document}